%% file: paper.tex
\pdfoutput=1
\documentclass[letterpaper, 10 pt, conference]{ieeeconf}
\IEEEoverridecommandlockouts                              % This command is only needed if 
\usepackage{times}

\usepackage{graphicx}
\usepackage{amsmath}
\usepackage{amssymb}
\usepackage{csquotes}
\usepackage{url}
\usepackage{hyperref}
\usepackage{subcaption}
\usepackage{color, colortbl}

\usepackage[]{svg}
\usepackage{multirow}
\usepackage{algorithm}% http://ctan.org/pkg/algorithms
\usepackage{algpseudocode} %http://ctan.org/pkg/

\usepackage[nolist]{acronym} % acronyms by the \ac{label} command
\let\labelindent\relax
\usepackage[inline]{enumitem} % inline enumerate
\usepackage{cite}
\usepackage{siunitx}
\newif\ifshowComments
\showCommentstrue
\newcommand{\os}[0]{CoBOS}

\input{macro.tex}

\title{CoBOS: Constraint-Based Online Scheduler for Human-Robot Collaboration}
\author{Marina Ionova$^1$ and Jan Kristof Behrens$^1$
\thanks{$^1$Czech Institute of Informatics, Robotics, and Cybernetics, Czech Technical University in Prague, \texttt{jan.kristof.behrens@cvut.cz}.}
\thanks{This work was co-funded by the European Union under the project ROBOPROX (reg. no. CZ.02.01.01/00/22\_008/0004590), by the AGIMUS project, funded by the European Union under GA no.101070165, and by the RICAIP project funded by the European Union's Horizon 2020 research and innovation program under grant agreement No.~857306. It also contributes to the sustainability of project CZ.02.1.01/0.0/0.0/16\_026/0008432 Cluster~4.0 -- Methodology of System Integration, financed by European Structural and Investment Funds and Operational Programme Research, Development and Education via Ministry of Education, Youth and Sports of the Czech Republic.}}

\begin{document}

\maketitle
% \thispagestyle{empty}
% \pagestyle{empty}

% \maketitle

\begin{abstract}
Assembly processes involving humans and robots are challenging scenarios because the individual activities and access to shared workspace have to be coordinated. Fixed robot programs leave no room to diverge from a fixed protocol. Working on such a process can be stressful for the user and lead to ineffective behavior or failure. We propose a novel approach of online constraint-based scheduling in a reactive execution control framework facilitating behavior trees called \os. This allows the robot to adapt to uncertain events such as delayed activity completions and activity selection (by the human). The user will experience less stress as the robotic coworkers adapt their behavior to best complement the human-selected activities to complete the common task. In addition to the improved working conditions, our algorithm leads to increased efficiency, even in highly uncertain scenarios. We evaluate our algorithm using a probabilistic simulation study with 56000 experiments. We outperform all other compared methods by a margin of $4-10\%$. Initial real robot experiments using a Franka Emika Panda robot and human tracking based on HTC Vive VR gloves look promising. 
\end{abstract}

% \IEEEpeerreviewmaketitle

\section{Introduction}

% \added{test} \deleted{test2}\replaced{new}{old}

Humans and robots working on a shared task promise to reap the best of two worlds: the robot's diligence and the human's dexterity and intelligence. However, robots are often perceived as slow, inefficient, and clumsy, i.e., bad colleagues. 
Thus, coordinating humans and robots on shared tasks more efficiently is a core concern (and a largely unsolved problem) of the growing collaborative robot (cobot) industry. Better coordination would lead to higher productivity, a reduced mental burden for humans in the presence of robots, and a better acceptance of robotic colleagues by the workers.
Task allocation in human-robot collaboration is, compared to the multi-robot collaboration from our previous work \cite{Behrens_Stepanova_Babuska_2020}, very challenging for many reasons. One is the uncertainty introduced by the human collaborator and amplified by complex task dependencies. The uncertainties comprise uncertainty regarding task durations, random machine breakdowns, or the rejection of tasks by the human worker. Furthermore, the knowledge of the system and its probability distributions will most certainly be wrong to some degree. 

\begin{figure}[!tb]
\begin{center}
\includegraphics[trim={0cm 2cm 0 0cm},clip,width= \linewidth]{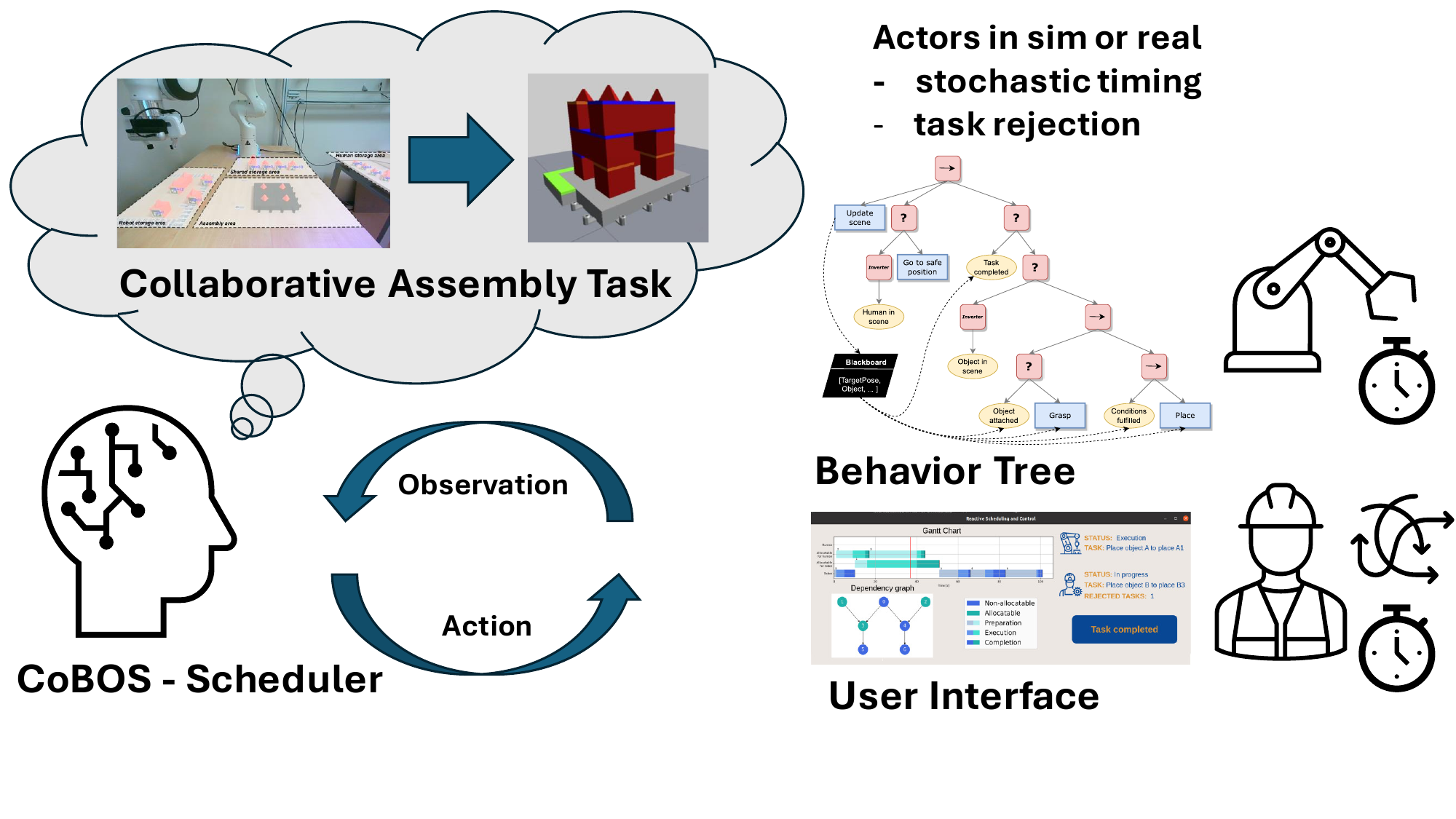}
\caption{\os{} is a constraint-based online scheduler for human-robot collaboration. It sequences tasks and allocates them to multiple actors based on uncertain information and incorporates observations during execution via event-driven rescheduling.}
\label{fig:real_robot_task_layout}
\end{center}
\end{figure}

% \begin{figure}[!tb]
% \begin{center}
% \includegraphics[trim={0cm 6cm 0 5cm},clip,width= \linewidth]{fig/Task_real_robot.jpg}
% \caption{The real robot setup and an example task based on the HRC task generator by \cite{Riedelbauch_Hümmer_2022}. The different production areas are annotated using the light gray overlay.}
% \label{fig:real_robot_task_layout}
% \end{center}
% \end{figure}

As a running example, we use the tabletop setup with a Franka Emika Panda manipulator shown in~\figref{fig:real_robot_task_layout}. A human worker can approach the shared workspace from the right side. A typical task is shown in~\figref{fig:real_robot_task_finished}. We use 3D printed assembly tasks as proposed by \cite{Riedelbauch_Hümmer_2022}. The workplace (see~\figref{fig:real_robot_task_layout} top left) is divided into an assembly area (the black construction plate), a part storage for the robot (left), a shared part storage (next to the robot base), and a part storage for the worker (right). The worker and the robot cannot access the assembly area and the shared part storage simultaneously. The human and the robot must collaborate to complete the assembly task (see the finished task state in \figref{fig:real_robot_task_finished}). Specifically, they have to coordinate their access to the shared areas and carry out the sub-tasks in an order consistent with the task's dependency graph shown in \figref{fig:real_robot_task_finished} on the right. 

Note that the sub-task allocation, ordering, and timing can be used to optimize the collaboration. These decisions significantly influence human-robot coordination, the robustness of the schedule, and finally, the makespan. 

This paper proposes \os{}, a Constraint Programming-based online scheduling method that we apply to and evaluate in the above-described Human-Robot Collaboration scenario. Our method gets a description of a decomposable task along with probability distributions and is informed of important events, such as sub-task completions, agent states, etc., during the execution. These observations are stated as facts in the model such that solving the model repeatedly will always yield a schedule that is consistent with what happened so far. While we want to achieve a short makespan ultimately, the shortest schedule might result in tight coupling of the agents' schedules so that unexpected outcomes might easily disrupt them to a degree that is more often than not sub-par to a more robust schedule. 
%Therefore, we incentive the planner to find an optimal trade-off between fast and robust schedules.

We evaluate our approach in an extensive probabilistic simulation study against two baseline methods and a reimplemented method from the literature\footnote{All simulation experiments in the paper can be reproduced using our released code \href{https://github.com/JKBehrens/CoBOS}{https://github.com/JKBehrens/CoBOS}}. The simulation environment allows comparing the performance of all methods in many runs sampled randomly from a given distribution. 
In total, we ran 56000 reproducible experiments that showed that our method outperformed the other methods. This advantage is more dominant in the more complex cases that require smart predictive decision-making.
%We also explore the effect of supplying wrong information to the method to see how well this mismatch is handled, as this will be the norm and not the exception in real-world situations. 

We summarize the contributions of this paper as follows: 
\begin{enumerate}
    \item A constraint-based online scheduler for human-robot collaboration called \os~that can schedule and execute tasks even when given wrong information about the task durations and in the presence of task rejections.
    \item A probabilistic simulation environment that can be used to thoroughly test multi-actor decision-making agents (such as \os) under uncertainty.
    \item The code and data required to reproduce the presented experiments.
\end{enumerate}

% This paper proposes a reactive architecture that utilizes Constraint Programming-based scheduling and Behavior Trees (BT) to tame the inherent uncertainty. We focus on the uncertainty in the timing of task execution and task rejection by humans. We assume that agents always finish their tasks and potential failures are repaired during an extended processing time. We evaluate our approach in an extensive probabilistic simulation study. Additionally, we implemented the method into a real system featuring a Franka Emika Panda robot, a vision system with marker-based task state tracking, and HTC Vive-based human activity tracking (see~Fig~\ref{fig:real_robot_task_layout}).

The remainder of the paper is structured as follows. In \secref{sec:related_work}, we discuss related works. ~\secref{sec:background} introduces preliminaries about Constraint Programming and Behavior trees that are required to understand the rest of the paper. In \secref{sec:reactive_architecture}, we discuss the class of use cases, the scheduling model, and the online scheduling architecture. \secref{sec:implementation} provides a short overview about the implementation. \secref{sec:experimental_setup} describes the conducted experiments and \secref{sec:results} discusses the achieved results. Finally, we conclude the paper in \secref{sec:conclusion}.

\section{Related Work}
\label{sec:related_work}

In this section, we provide an overview of the related work regarding online scheduling for Human-Robot Collaboration.

\etal{Ghaleb} \cite{Ghaleb_Zolfagharinia_Taghipour_2020} report on a computational study about Flexible Job Shop Scheduling. Specifically, they compared how and when to schedule and reschedule. They identified event-triggered rescheduling as particularly effective. \os~follows the same strategy. However, for our HRC problems, we did not encounter the issue that the computations would be too slow to outperform heuristic solutions. They also do not specifically deal with HRC and human-specific modes of uncertainty.

An interesting approach to endow autonomous systems with faster and better reactions is to provide them with an \textit{executive} that has certain reactions and decisions already embedded. Drake is such a system for Temporal Plans with Choice \cite{Conrad_Williams_2011}. However, they deal only with single actors. Compiling an executive from our schedules could be considered for future work in combination with proactively exploring potential threats to the schedules \cite{Cyras_Letsios_Misener_Toni_2019}. We use Behavior Trees \cite{Colledanchise_Ögren_2018} as executive but we do not generate them entirely from plans.

A method for dynamically allocating (DA) tasks in HRC scenarios was proposed by \etal{Petzoldt} \cite{Petzoldt_Niermann_Maack_Sontopski_Vur_Freitag_2022}. We compare \os~against our implementation of DA and find that DA beats all baseline methods, but it cannot exploit the decision freedom in the harder problems as efficiently as \os. This might be because DA does not look ahead to the end of the schedule and might miss options for load balancing etc.

A new direction of scheduling multi-agent systems is Reinforcement learning-based scheduling \cite{Zhou_Tang_Zhu_Zhang_2021}, \cite{Zhang_Lv_Bao_Zheng_2023}. While learning might produce powerful solver or surrogate functions for IoT or Industry 4.0 settings, in HRC data is sparse, and the uncertainty is higher. Also, the latency requirements are more strict. However, we plan to enable \os~in the future to learn probability distributions or schedule grading functions. 

Another recent work, the PLATINUM system, utilizes timeline-based scheduling \cite{Umbrico_Cesta_Cialdea_Mayer_Orlandini_2017}. While PLATINUM implements its flaw-resolver-based temporal planning with uncertainty, we utilize similar data structures in our model. Still, we reduce the problem to Constraint Satisfaction Problems and use an off-the-shelf solver that produces solutions fast enough to run online. Most often, the proposed scheduling model proves even the optimality of the returned solution in fractions of a second.

\section{Preliminaries and Background}\label{sec:background}
In this section, we introduce concepts about constraint programming-based scheduling and reactive behavior generation using behavior trees that are required to understand the content of this paper.

\subsection{Constraint Programming for Scheduling}

In Constraint Programming (CP), we model a (planning) problem declaratively in terms of Constraint Satisfaction Problems (CSP).
A Constraint Satisfaction Problem is generally specified by a triple $\mathcal{P} = (X,D,C)$, where $X$ is a $n$-tuple of variables $X = \{x_1, x_2, \ldots, x_n\}$, 
$D$ is a $n$-tuple of domains $D = \{D_1, D_2, \ldots, D_n\}$, and $C$ is a $t$-tuple of constraints $C = \{C_1, C_2, \ldots, C_t\}$. The domain $D_i$ maps the variable $x_i$ to possible values of $x_i$: $D(x_i) = D_i$, i.e., $x_i \in D_i$. A constraint $C_j$ is a tuple $\{R_{S_j}, S_j\}$, where $S_j$ is the subset of variables in $X$, which are involved in the constraint $C_j$. $R_{S_j}$ is a relation between the variables $S_j$, which effectively defines a subset of the Cartesian product of the domains of the variables in $S_j$ \cite{freuder_onstraint_2006}. A solution of a CSP is a complete assignment $A = \{a_1, \ldots, a_n\}$, which assigns to each variable $x_i \in X$ a value $a_i$, which is within the domain $D(x_i)$ of this variable:% of values from the variable domain to variables $X = \{x_1,..., x_n\}$: 

$$x_i \mapsto a_i, \forall i \in \{1,\ldots,n\} \mathrm{, where~} a_i \in D(x_i).$$ 

%i.e., to each variable $x_i$ we assign a value $a_i$, which is within the domain of this variable. 
The task of a constraint solver is to find such a set $A$, such that $A$ satisfies all constraints $C$. For minimization, the solver shall find the $A \in \mathcal{A}$ that minimizes or maximizes the value of a designated variable. $\mathcal{A}$ is the space of all solutions. 

Note, that the CP formulation can be used for finding solutions, verifying solutions, completing partial solutions, and analyzing hypothetical scenarios (i.e., what-if analysis).

\subsection{Behavior Trees}

Behavior trees (BT) are a behavior modeling tool that is inherently composable and reactive. 
BTs were initially developed for the game industry to control Non-Player-Characters but recently received much attention in robotics deliberation \cite{Colledanchise_Ögren_2018, Iovino_Scukins_Styrud_Ögren_Smith_2022}. BTs could very well be the link between long-running deliberation methods, such as AI planning, and robotic acting that needs to deal with uncertainty and react to external perturbations of the environment \cite{Martín_Morelli_Espinoza_Lera_Matellán_2021}.

BTs are trees composed of behavior nodes, processors, and modifiers. Leaf nodes are actual behaviors or condition checks. Intermediate nodes are processors or modifiers. 
The central operation in a BT is the \textit{tick}. %
%The intermediate nodes are *processors* and the leave nodes are *primitives* or *conditions*. 
The root node is regularly \textit{ticked}. Each node recursively ticks its child nodes and returns a status of either \textit{running}, \textit{success}, or \textit{failure}. The processors decide how to handle the states of their children. Typical processor nodes are \textit{sequence}, \textit{parallel}, \textit{selector}, but many variants exist. 

BTs can be, due to their clear structure, manipulated programmatically. This also allows us to refine them on the go. We could add a node for an abstract action into the tree and expand it when the node is ticked (for example, by planning). In this paper, we encode a base robotic behavior as BT and encode the allocated tasks as data in the robot's blackboard. 

\section{Online Scheduling Architecture} \label{sec:reactive_architecture}

In this section, we present the details about the use case formalization, the CP formulation for scheduling, and the integration into an online deliberation for HRC.

\subsection{Use Case Specification}

In this work, we consider shared manufacturing jobs that can be decomposed into individual pick\&place like sub-tasks. Each sub-task has to be allocated to an actor (e.g., the human or the robot) capable of executing the task. Here, this depends on the initial placement of the handled part. Each actor can only work on a single task at a time.

\begin{figure}[!hptb]
\begin{center}
\includegraphics[width= \linewidth]{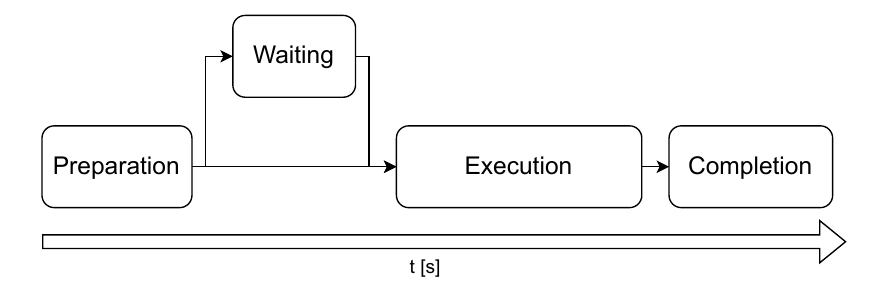}
\caption{Task graph of each sub-task consisting of the three phases \textit{preparation}, \textit{execution}, and \textit{completion}. An optional \textit{waiting} phase is allowed between preparation and execution to accommodate for an occupied shared area or unmet task dependencies.}
\label{fig:task-intervals}
\end{center}
\end{figure}

Without loss of generality, we model each task as three phases: 1) preparation--the actor picks an object and moves towards the central assembly area, 2) execution--the actor enters the assembly area, places the part, and leaves the area, and 3) completion--the actor moves home. Each phase has a duration that is dependent on the actual work done, the state of the environment, and the actor. This is effectively modeled as a mixture of Gaussian distributions where the different components of the mixture approximate the distribution of normal executions and failure cases.

\begin{figure}[!hptb]
\begin{center}
\includegraphics[trim={0cm 10cm 0 5cm},clip, width= 0.8\linewidth]{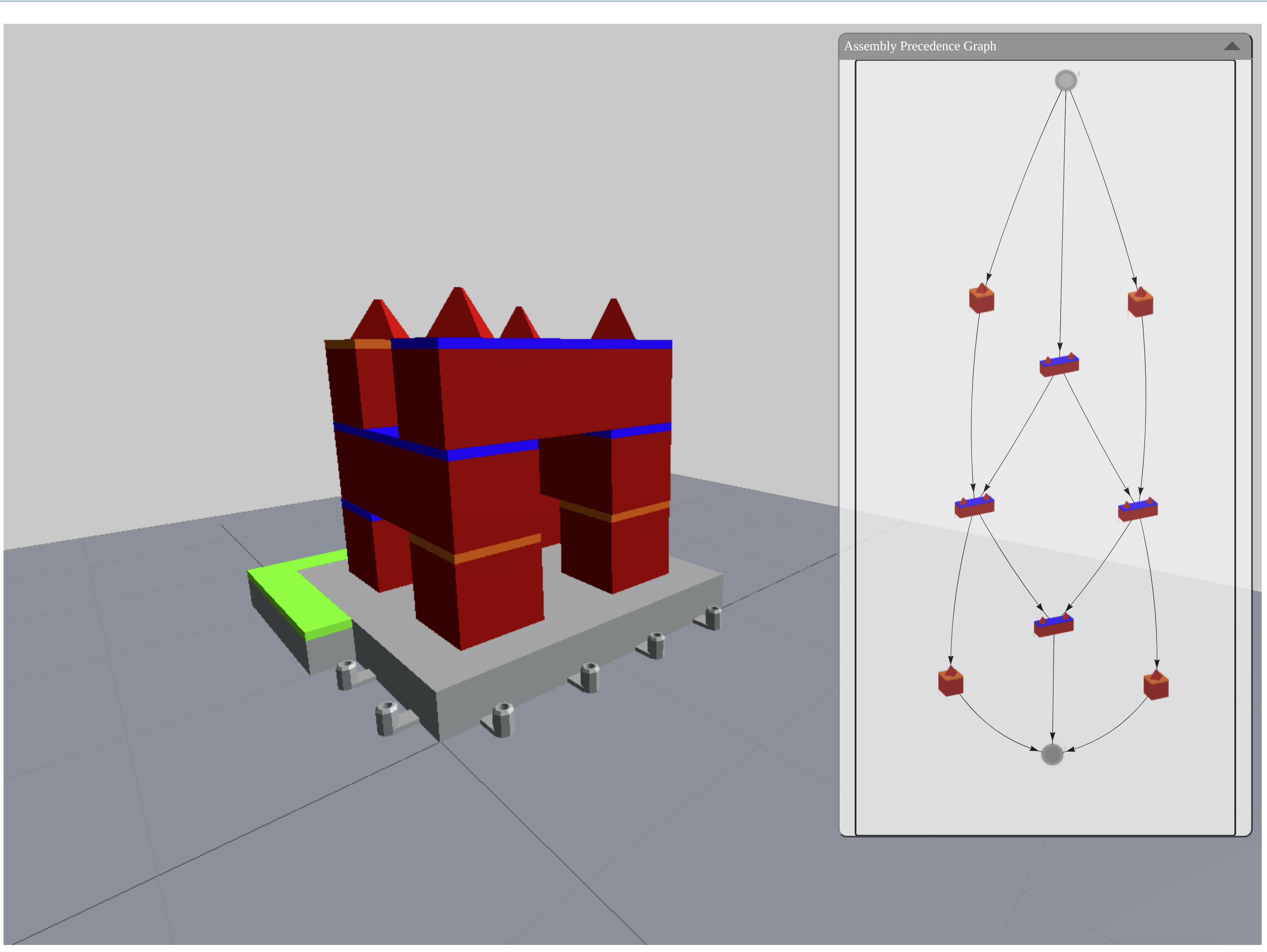}
\caption{The final state of the shared assembly task. The precedence graph is shown on the right.}
\label{fig:real_robot_task_finished}
\end{center}
\end{figure}

Sub-tasks can be dependent on the fulfillment of other tasks (i.e., precedence constraints). \figref{fig:real_robot_task_finished} shows a typical task and its dependency graph. The \textit{execution} phase of a task can only start when the \textit{execution} phases of all its dependencies are completed. Therefore, a \textit{waiting}-phase between the preparation and the execution is allowed.

Finally, the actors must avoid collisions. Therefore, the actors cannot access shared areas simultaneously but use the \textit{waiting} phase until entering is possible.

% \jb{these are the rules of the env}
% \jb{integrated planning and acting system}
% \jb{job scheduling subsection}

% \begin{figure}[!h]
% \begin{center}
% \includesvg[inkscapelatex=false, width= \linewidth]{fig/gantt_chart.svg}
% \caption{\jb{add picture showing task dependencies } }
% \label{fig:real_robot_task_dependency_graph}
% \end{center}
% \end{figure}

\subsection{Probabilistic Collaboration Simulation}
\label{sec:prob-collab-sim}

\begin{figure}[tbh]
\begin{center}
\includegraphics[width=\linewidth]{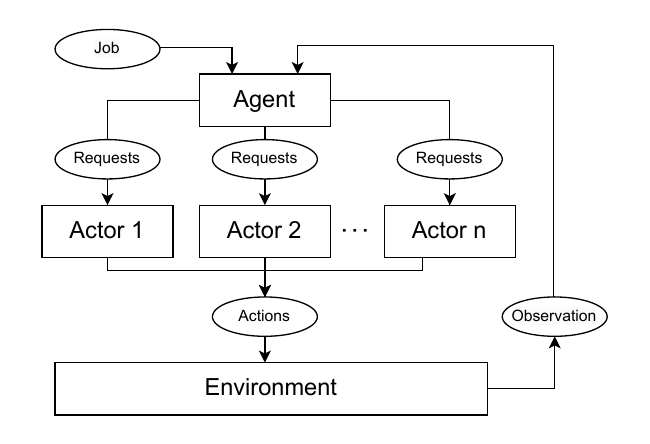}
\caption{Our proposed probabilistic closed-loop multi-actor simulation is a test environment template for central multi-actor coordination agents. The agent, for example \os{}, requests, based on a job description and continuous observations, actions from the actors. The actors, e.g., a human worker or a robot, act upon these according to their policy. Humans, for example, might reject a task, while robots generally will accept assigned tasks. The environment decides for values of uncontrollable variables and compiles the observation message at each time step. All components are deterministic for a given random seed.}
\label{fig:prob_sim_loop}
\end{center}
\end{figure}

The proposed simulation environment is shown in \figref{fig:prob_sim_loop}.
To simulate collaboration, we divide the acting environment into an agent $s$ and several actors $a_i$. The agent (e.g., our scheduling-based approach) makes decisions and communicates them to the actors. These messages resemble the prompt on the user's screen or the added task to the agenda of the robot. The actors decide according to an actor policy $\pi_i$ about all controllable variables $u_a^i$. The actors are subject to uncertainty controlled by the simulation, i.e., the environment decides about the uncontrollable variables $y_a^i$. When applying the actors' actions, the environment also ensures that no two actors are accessing a shared area simultaneously. The environment generates observation data at each time step that contains the state of each actor (e.g., \textit{idle}, \textit{waiting}, \textit{execution}), the state of each task (e.g., \textit{unavailable}, \textit{available}, \textit{inprogress}, \textit{completed}), and a set of additional task constraints (e.g., tasks rejected by the human, other observations). The simulation determines task durations and other probabilistic states (e.g., the task acceptance decision) by sampling a priory from the underlying distributions. However, the observation will reveal this data only at the appropriate time to the agent. 

% \jb{this is moved here from the exp section. Integrate...}
The task execution time is sampled from a bimodal probability distribution, where one mode represents normal execution, and the other represents failed attempts, which leads to increased task duration. We used the simulation to generate test data for our method implementations (e.g., assignments were allocated and sequenced correctly).
Furthermore, the rescheduling was tested to be consistent with all observed events. 
The simulation was designed to accept the signals the actors would receive about the start of the tasks. Uncontrollable events, such as the completion or the rejection of a task, are controlled by the simulation.  
Specifically, the scheduling algorithm assigns tasks to robots (and simulated humans). Then, after sufficient amount of simulation time, the simulated task is complete, and the feedback to the algorithm about the completion of the task, including information about the time for each phase, is sent.

In addition to the temporal uncertainty we also simulate the human choice. When there is a decision to be made about accepting a task or a preference, the simulation randomly samples from a probabilistic distribution for that task. For instance, if the refusal rate for a task is 30\%, and the robot offers to perform the task, the human participant in the simulation will accept the task with a probability of 70\%. This feature allows the simulation to test the scheduling algorithm's ability to handle human decision-making uncertainties.

% % \jb{how processed (symbolic vs. sub-symbolic is the observation) }
% \begin{figure}[tb]
% % \begin{algorithm}
% \begin{algorithmic}[1]
% \State \textbf{Inputs:} job description $\mathbf{P}$, agent $S$, set of actors $\mathcal{A}$, 

% \State $t \gets 0$
% \State sim.init()

% \While{$P.progress() < 100$}
%     \State $O \gets$ sim.getObservation()
%     \State Requests $\gets$ S.decide($O$)

%     \For{$a \in \mathcal{A}$}
%         \If{$a \in$ Requests $\bigcap$ $a$.isIdle()} 
%             \State state $\gets a$.StartTask()
%         \Else
%             \State state $\gets$ a.proceed()
%         \EndIf

%     \EndFor
% \EndWhile
% \end{algorithmic}
% % \end{algorithm}
% \caption{\jb{it would be nice to have a concise alg of what the sim loop is doing. }}
% \label{alg:prob-scheduling-simulation}
% \end{figure}

\subsection{Integrated Planning and Acting System}

\begin{figure}
    \centering
    \includegraphics[width=\linewidth]{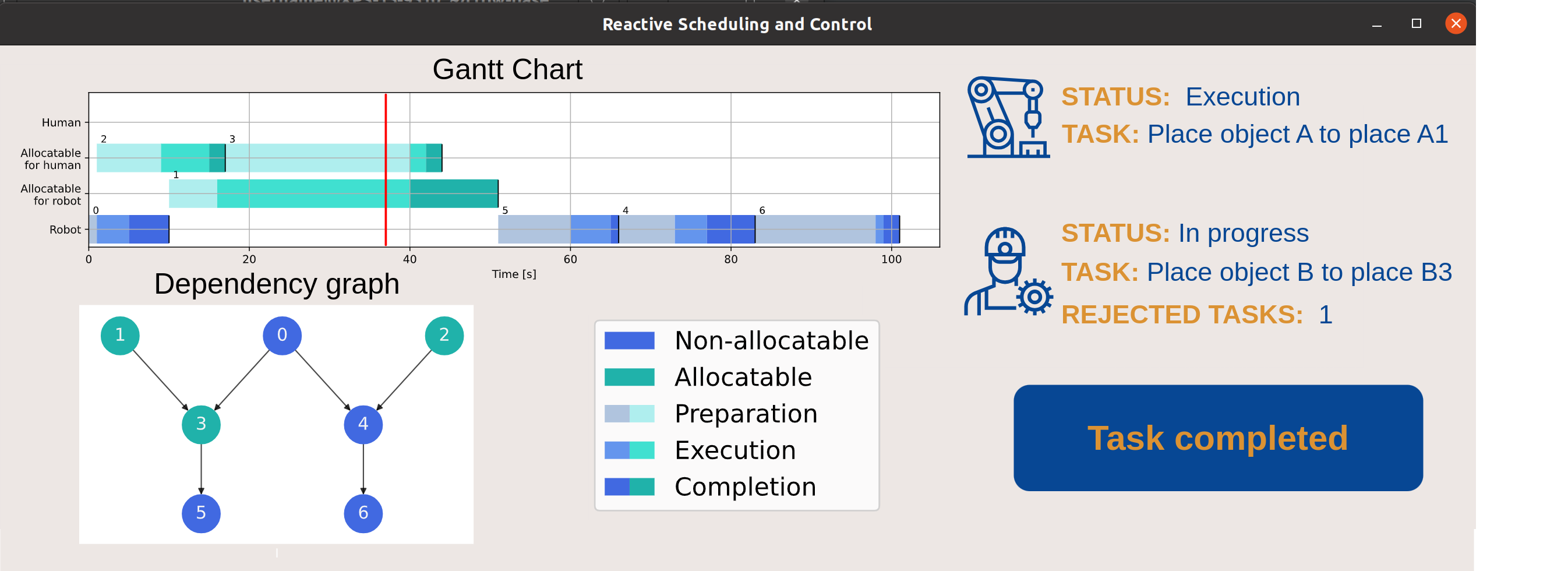}
    \caption{Graphical User Interface for the real setup. A dynamic Gantt chart (top-left) shows the current state and schedule of the task. The dependency graph (bottom-left) shows the task-specific dependencies. On the right side, the current task details are displayed for the robot and the user. Human feedback about task completions or task rejections is collected via buttons.}
    \label{fig:GUI}
\end{figure}

The online execution requires the management of the robotic and human actors, continuous supervision, and the incorporation of new observations about the tasks' progress into the scheduling. 
This subsection describes the decision-making agent shown in \figref{fig:prob_sim_loop} at the top. It receives as input a job description and, at each timestep, an observation. It then determines actions for each \textit{idle} actor. These are sent as \textit{requests} to the actors. Here, requests to the human actor are sent via the GUI shown in \figref{fig:GUI}. Requests to the robot(s) are written to the respective behavior tree blackboard (see~\figref{fig:bt_hrc}). 

\subsubsection{Graphical User Interface}

The human worker is informed about the task state using a GUI. Specifically, a dynamic Gantt chart (top-left) shows the current state and schedule of the task, and the dependency graph (bottom-left) shows the task-specific dependencies. On the right side, the current task details, and the task statuses, are displayed for the robot and the user. In addition, it presents a list of tasks rejected by the human worker. Human feedback about task completions or task rejections is collected via buttons in the main or pop-up window.

\subsubsection{Behavior tree robot control}
\begin{figure}[!htb]
% \begin{center}
\centering
\includegraphics[width=\linewidth]{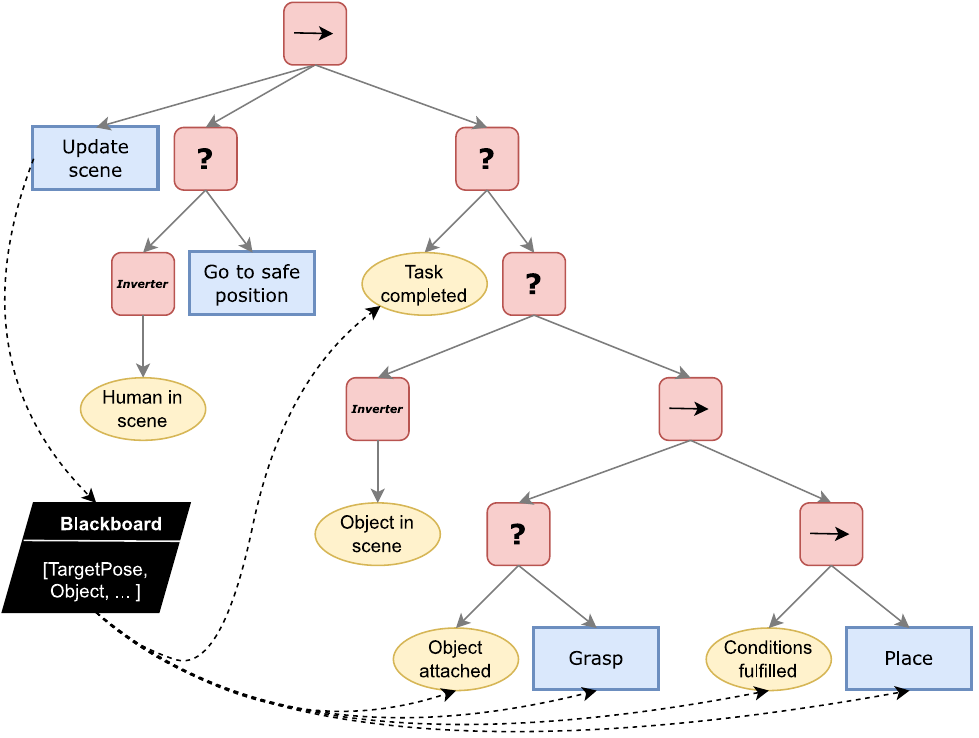}
\caption{Behavior tree for managing the robotic agent. Successively, all tasks are written to the blackboard when they are due according to the schedule. First-added tasks have priority. In this way, the robot will repair the situation if accidentally progress is undone. 
%\added{Full version of BT: https://drive.google.com/file/d/1GEjl9mLO51gTjRb2dHeLiOM3k4RjSQS5/view?usp=sharing}
}
\label{fig:bt_hrc}
% \end{center}
\end{figure}

The robot is controlled by the BT shown in \figref{fig:bt_hrc}, which ensures that the robot evades the human worker with high priority, works on the given subtasks with middle priority, or returns to its home pose when idle.

% \jb{marina, in \figref{fig:bt_hrc} please remove the unnecessary going home fallback. remove also the repeat on top.}

\subsubsection{Job Scheduling Model}

% The scheduling model is always run when no schedule was computed or new conflicting evidence makes the last schedule obsolete (e.g., end time points or task rejections). Observed facts are constrained to their actual values because the solver should not reason about the past.
In this section, we detail the constraint model that is used to find schedules. The objective is to minimize the makespan $m$.
\begin{equation}
\begin{aligned}
& \underset{\mathbf{x}\in \mathcal{A}}{\text{minimize}}
& & m(\mathbf{x})\mathrm{,} \\
% & \text{subject to}
% & & f_i(x) \leq b_i, \; i = 1, \ldots, m.
\end{aligned}
\end{equation}
where $\mathbf{x}$ is an element of the set of valid solutions $\mathcal{A}$ to the CSP made of the following variables and constraints. 

The formulation of the scheduling model builds on the notion of time points and intervals. Let $t_{ips}$ and $t_{ipe}$ be the start and end of the $p$-th phase of the $i$-th task, respectively. The duration of phase $p$ task $i$ is $d_{ip}$, and for all intervals holds the constraint 
\begin{equation}
t_{ipe} = t_{ips} + d_{ip} \mathrm{.}    
\end{equation}
Our task model has three phases. All phases follow right after each other but only the durations of phases $2$ and $3$ are fixed. Phase $1$ can accommodate the optional waiting phase.
\begin{equation}
\forall i \in T:  t_{i3s} = t_{i2e} 
\end{equation}
\begin{equation}
\forall i \in T: t_{i2s} = t_{i1e} 
\end{equation}
\begin{equation}
\forall i \in T, \forall p \in  \{2,3\}:  d_{ip} = \hat{d}_{ip} 
\label{eq:phase-dur23}
\end{equation}
$\hat{d}_{ip}$ is a given estimate of the $d_{ip}$.
\begin{equation}
\forall i \in T: d_{i1} \geq \hat{d}_{i1}
\label{eq:phase-dur1}
\end{equation}
We define a variable $m$ and constrain it to take the value of the makespan, which we use as a minimization target.
\begin{equation}
m = \underset{i \in T}{\text{max}}~t_{i3e}   
\end{equation}
The sub-task dependency graph $\mathcal{G} = (T, E)$, where $T$ is the set of all sub-tasks and $E$ is the set of all edges representing precedence constraints. A valid dependency graph $\mathcal{G}$ must be directed and acyclic. If an edge $e \in E$ connects a vertex $i$ with vertex $j$, then task $i$ depends on $j$. Task dependencies imply temporal ordering among the execution phases of the involved tasks. 
\begin{equation}
 \forall (i, j) \in E: t_{i2s} \geq t_{j2e}
\end{equation}
$T$ is the set of all sub-tasks.
Each task $i \in T$ needs to be assigned to an actor. The variables $x_{ia}$ indicate that task $i$ is assigned to actor $a$.
\begin{equation}
\forall i \in T: \sum_{a \in A} x_{ia} = 1  
\end{equation}
Each actor can only work on a single task at a time. Hence, sub-tasks allocated to an actor may not overlap.
\begin{equation}
 \forall i,j \in T | i \neq j: x_{ia} = x_{ja} \implies t_{is} > t_{je} \bigvee t_{js} > t_{ie}    
\end{equation}
Note, that we encode these constraints for efficiency reasons using a single \textit{NoOverlap} global constraint on optional interval variables.
For each shared resource, i.e., the shared areas, we enumerate the intervals and include them in a \textit{NoOverlap} constraint.

\subsubsection{Online model management}

The model explained in the previous section can be used to calculate a schedule with the exact timing for each task. However, to incorporate online observations of uncertain events, we must change the model to prevent conflicting constraints and infeasible models. 
Here, we deal with four cases of observations.
\begin{enumerate}
    \item When actor $a$ starts task $i$, we fix $t_{i1s}$ and $x_{ia} = 1$.
    \item When an actor finishes a task or task phase. We remove the constraints on the durations of the tasks \equationref{eq:phase-dur1} and \equationref{eq:phase-dur23} and add the observed values to the respective time variables $t_{ipe}$. If the task is not finished within the allotted time, we extend the phase to the following time step, i.e., $t_{ipe} = t_\mathrm{now} + 1$. 
    \item When an actor $a$ rejects an assigned task $i$, we add a constraint $x_{ia} = 0$.
    \item For any future task, we set the domain of its starting time to  $D(t_{i1s}) = \{now \ldots t_H\}$, where $t_H$ is a maximum horizon. This ensures that no task is planned for the past.
\end{enumerate}
These changes allow us to compute new schedules consistent with the current observation state, even if the observations conflict with previous assumptions, e.g., about the task durations.

\section{Implementation} \label{sec:implementation}

Our implementation is based on ROS 2. Our BTs utilize the Pytrees library. Object detection and 6D localization use  Intel Realsense cameras and fiducial markers on the objects. The human pose is tracked using an HTC Vive tracker. The robot is controlled using position-based servoing \cite{rtb} using a joint impedance controller of panda-py \cite{elsner2023taming}. The scheduling is implemented using Google OR-Tools' CP-SAT solver \cite{google_2024}. 
To make the battery of simulation experiments, we utilized the Dask Python library \cite{rocklin2015dask} to parallelize our code and deploy it to our High-Performance Compute cluster.

\section{Experimental Setup}\label{sec:experimental_setup}

We designed a probabilistic simulation, introduced in \secref{sec:prob-collab-sim}, to evaluate the effectiveness of the scheduling method under uncertainties in task rejection by humans and the task duration. The purpose of the simulation is to emulate the behavior of the collaborating human and the environment. The simulation is transparent to the decision-making approach because it offers the same interface as the real environment. The scheduler commands the real or simulated robot and indicates task assignments to the real or the simulated worker. However, the simulation probabilistically decides the task durations and whether the worker accepts or rejects assigned tasks. 

Inspired by the use cases that can be generated using the HRC task generator \cite{Riedelbauch_Hümmer_2022} and the multi-robot task allocation (MRTA) taxonomy \cite{Korsah_Stentz_Dias_2013}, we designed a set of seven tasks of increasing difficulty. 

The tasks are as follows:
\begin{itemize}
    \item Task 1: Simple pick and place operation. The robot and the human have fixed tasks assigned, i.e., they pick up an object from a specified location and place it at designated spots. The decisions concern only the ordering of tasks and their exact timing. Timing uncertainties are present. This task contains no dependencies (ND).
    \item Task 2: Same as Task 1 but with the additional freedom to allocate a subset of the tasks freely. This can be used to balance the workload and thus reduce the makespan. The human worker can reject allocatable tasks. However, this also leads to complex dependencies (CD) because multiple task decompositions exist.
    \item Task 3: Collaborative assembly. The robot and the human worker need to coordinate their work together to assemble a set of simple structures. The tasks concerning each structure are subject to ordering constraints. No tasks are allocatable; hence, cross-schedule dependencies arise (XD).
    \item Task 4: Same as Task 3 but with allocatable tasks. This makes the dependencies complex (CD) and good schedules harder to establish.
    \item Task 5: Complex assembly with more parts and dependencies. No task allocation (CD).
    \item Task 6: Same as task 5 but with allocatable tasks, which leads to complex dependencies (CD).
    \item Task 7: Randomly generated tasks with more tasks, generated dependency graphs, and task allocation options (CD). 
    
\end{itemize}

Each task is designed to test if the decision-making can exploit the structure of the task and how much the uncertainty impacts the resulting makespan.

We devise four methods: \os{} (ours), two greedy baseline methods Random Allocation (RA) and Maximum Duration (MD), and Dynamic Allocation (DA)~\cite{Petzoldt_Niermann_Maack_Sontopski_Vur_Freitag_2022} from the literature. Each method takes the generated observations and devises actions for each actor, i.e., it takes the place of the agent in \figref{fig:prob_sim_loop}. 
We generated ten instances for each case and tried each method 100 times on each instance, with task rejection and 100 times without. Each time the method got a different estimate for the task durations and rejection probabilities that were sampled from the correct underlying distributions. We also solved each instance to optimality using our solver assuming perfect information. This strong theoretical lower-bound on the makespan allows us to normalize the resulting makespans and compare all $1000$ runs on the same case class.

\begin{figure}[htb]
% \begin{center}
\centering
% \includesvg[inkscapelatex=false, width=0.8\linewidth]{fig/bimodal_prob.svg}
\includegraphics[width=\linewidth]{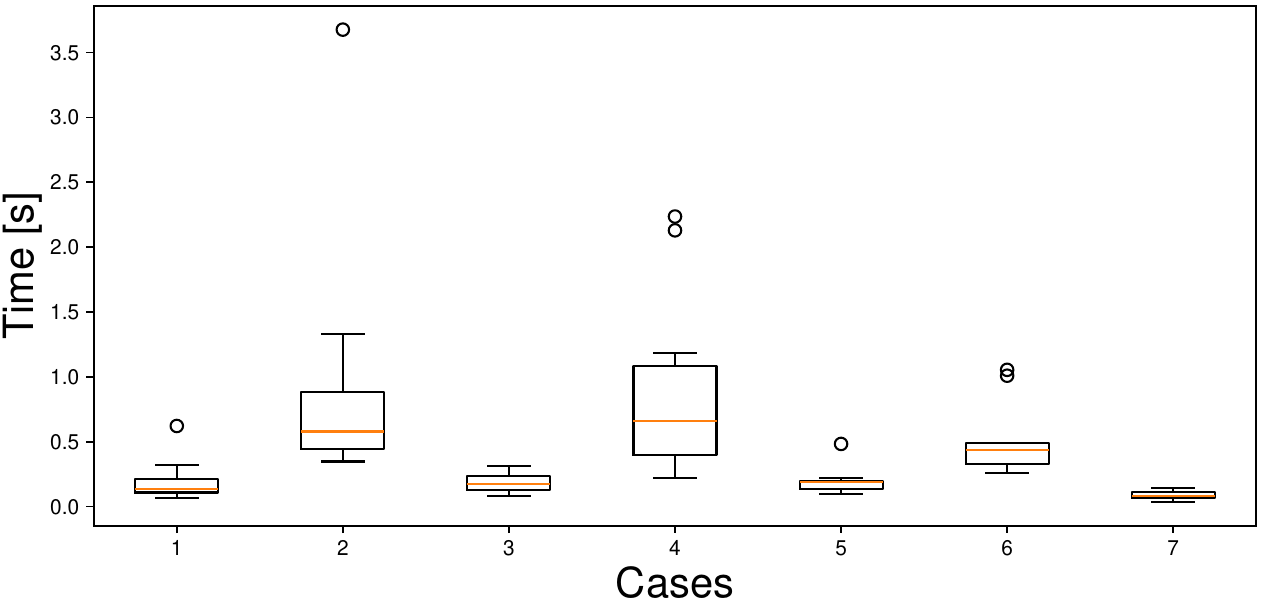}
\caption{This graph illustrates the initial decision-making duration of CoBOS, including the creation of the CP model, solving for the schedule, and assigning tasks to agents. As tasks are completed and variables become fixed, subsequent decisions take less time. The data indicates that the mean and standard deviation of calculation time are higher in cases involving allocatable tasks. Other methods are not shown in the figure because their decision-making time is negligible due to instantaneous assignment (IA), whereas our approach involves a time-extended assignment (TA)~\cite{Korsah_Stentz_Dias_2013}.}
% \caption{This graph illustrates the initial decision-making duration of CoBOS, encompassing the processes of creating a CP model, solving the schedule, and assigning tasks to individual agents. Subsequent decisions required less time due to the fixed variables resulting from completed tasks. The data reveals that the mean and standard deviation of calculation time are greater in cases involving allocatable tasks.}
\label{fig:task-duration-gmm}
% \end{center}
\end{figure}

% In addition to considering the completion times of individual tasks, the simulation also considered the dependencies between tasks. This was done by calculating the wait times required for each task and ensuring that it did not start earlier than it should have based on these dependencies. By considering these dependencies and wait times, the simulation could accurately evaluate the effectiveness of the scheduling algorithm and ensure that it was working correctly.

\section{Experimental Results}\label{sec:results}
\begin{figure}[!tb]%htbp]
    \centering
    % \includesvg[width= \linewidth]{fig/boxplot_case_8_grid.svg}
    \includegraphics[width=\linewidth]{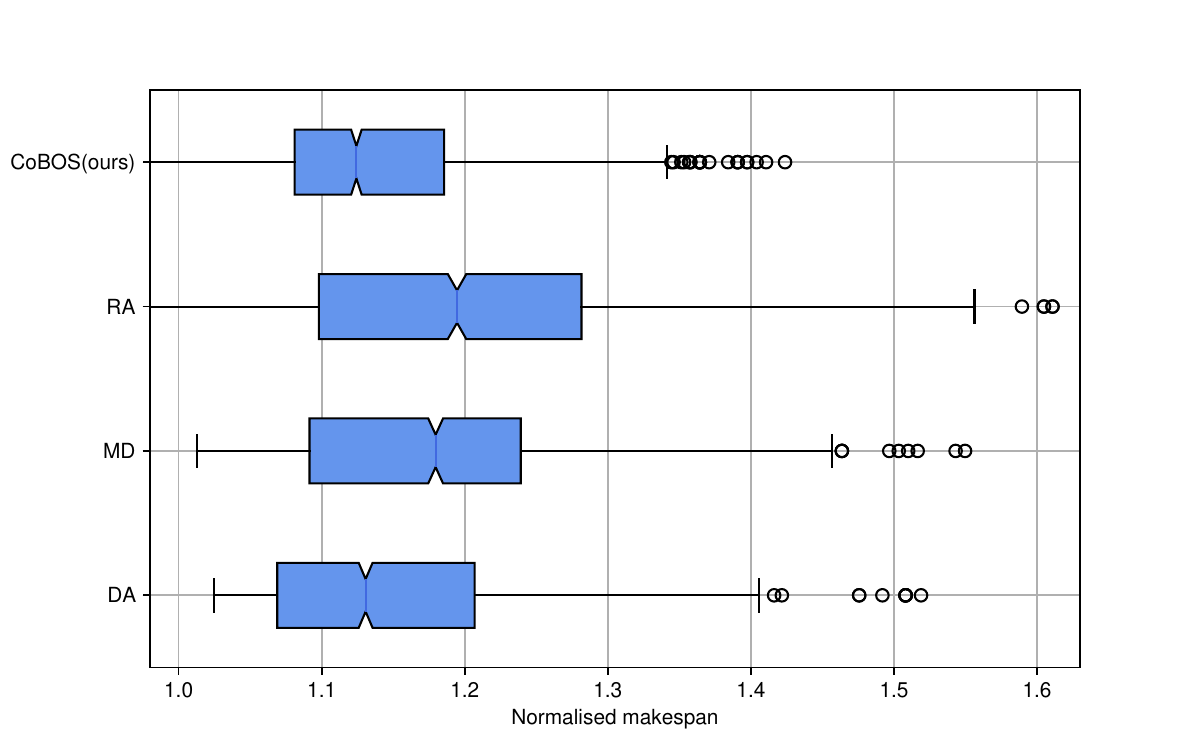}
    \caption{The graph depicts the normalized makespan distribution (smaller is better) within the experiment conducted for case 7, i.e., randomly generated jobs. The methods employed in the experiments are presented on the vertical axis. Our proposed approach outperforms the other methods considerably. It achieves lower average makespans as well as a lower variance. This indicates that our method effectively utilizes task flexibility and adeptly manages scenarios with increased uncertainty.}
    \label{fig:makespan_scenarios}
\end{figure}

We performed 56000 simulation experiments with different task allocation distributions, probabilities of human task rejection, and probabilities of error for each agent to test the efficiency and robustness of the proposed algorithm with varying task dependencies. 
% The time performance of our method shows lower computational times in more complex cases (see~\figref{fig:task-duration-gmm}). Cases with fewer dependencies or no allocatable tasks exhibit higher calculation times.
% The results were compared with the baseline Dynamic Allocation (DA)\cite{Petzoldt_Niermann_Maack_Sontopski_Vur_Freitag_2022} and two greedy methods Random Allocation (RA) and MaxDuration (MD). 
\tabref{tab:result_table} reports the performance of each method relative to the theoretical lower bound. It can be seen that \os~outperforms the other methods by a margin dependent on the case's complexity. In all cases, the variance of our method is lower, and thus our method is more reliable in finding good solutions. \figref{fig:makespan_scenarios} visualizes this for case 7, i.e., the most complex scenario.
\figref{fig:task-duration-gmm} shows the decision-making time for each scenario. Case 7, a realistic and complex scenario, had shorter computational times compared to the test cases, likely due to less symmetry and more constraints, which reduced decision options. This suggests promising performance for real-world problems.
% \figref{fig:task-duration-gmm} shows the decision-making time for each scenario. Case 7, a complex and realistic case, had relatively short computational times compared to the test cases, indicating promising performance for real-world problems. 

%The considered scenarios are fixed task allocation, free task allocation with the option of task rejection for humans, and free task allocation without task rejections. 

% \begin{figure}
%     \centering
%     \includegraphics[width= \linewidth]{fig/histogram_pd.png}
%     \caption{\jb{description}}
%     \label{fig:perf_cases}
% \end{figure}

% \multirow{4}{*}{\rotatebox{90}{$\mu$}}

% Case 7 in this table is our case 8. I just wanted to avoid questions about the missing case.
% case columns have been changed to fit the description
% original order 1 3 5 2 4 6
\begin{table}[htbp]
\resizebox{\linewidth}{!}{
\begin{tabular}{|cc|l|l|l|l|l|l|l|}
\hline
\multicolumn{2}{|c|}{Case}                                                              & \multicolumn{1}{c|}{1} & \multicolumn{1}{c|}{2} & \multicolumn{1}{c|}{3} & \multicolumn{1}{c|}{4} & \multicolumn{1}{c|}{5} & \multicolumn{1}{c|}{6} & \multicolumn{1}{c|}{7} \\ \hline
\multicolumn{1}{|c|}{\multirow{4}{*}{\rotatebox{90}{$\mu$}}}    & \os & \textbf{1.05}          & \textbf{1.15}          & \textbf{1.05}          & \textbf{1.15}          & \textbf{1.05}          & \textbf{1.10}           & \textbf{1.13}          \\ \cline{2-9} 
\multicolumn{1}{|c|}{}                                                             & RA & \textbf{1.05}          & 1.19                   & 1.08                   & 1.19                   & 1.11                   & 1.18                   & 1.22                   \\ \cline{2-9} 
\multicolumn{1}{|c|}{}                                                             & MD & \textbf{1.05}          & 1.19                   & 1.08                   & 1.19                   & 1.11                   & 1.19                   & 1.20                    \\ \cline{2-9} 
\multicolumn{1}{|c|}{}                                                             & DA & \textbf{1.05}          & \textbf{1.15}                   & 1.08                   & \textbf{1.15}                   & 1.11                   & 1.15                   & 1.17                   \\ \hline
\multicolumn{1}{|c|}{\multirow{4}{*}{\rotatebox{90}{10\%}}}     & \os & \textbf{1.0}           & \textbf{1.07}          & \textbf{1.0}           & \textbf{1.07}          & \textbf{1.00}           & \textbf{1.02}          & \textbf{1.04}          \\ \cline{2-9} 
\multicolumn{1}{|c|}{}                                                             & RA & 1.0                    & 1.08                   & \textbf{1.0}           & 1.08                   & 1.01                   & 1.07                   & 1.09                   \\ \cline{2-9} 
\multicolumn{1}{|c|}{}                                                             & MD & 1.0                    & 1.08                   & \textbf{1.0}           & 1.08                   & 1.03                   & 1.07                   & 1.08                   \\ \cline{2-9} 
\multicolumn{1}{|c|}{}                                                             & DA & 1.0                    & \textbf{1.07}          & \textbf{1.0}           & \textbf{1.07}          & 1.03                   & 1.04                   & 1.07                   \\ \hline
\multicolumn{1}{|c|}{\multirow{4}{*}{\rotatebox{90}{90\%}}}     & \os & \textbf{1.10}           & \textbf{1.25}          & \textbf{1.1}           & \textbf{1.25}          & \textbf{1.11}          & \textbf{1.21}          & \textbf{1.25}          \\ \cline{2-9} 
\multicolumn{1}{|c|}{}                                                             & RA & 1.11                   & 1.30                    & 1.2                    & 1.30                    & 1.21                   & 1.31                   & 1.37                   \\ \cline{2-9} 
\multicolumn{1}{|c|}{}                                                             & MD & 1.11                   & 1.31                   & 1.2                    & 1.31                   & 1.21                   & 1.36                   & 1.32                   \\ \cline{2-9} 
\multicolumn{1}{|c|}{}                                                             & DA & 1.11                   & 1.26                   & 1.2                    & 1.26                   & 1.21                   & 1.27                   & 1.29                   \\ \hline
\multicolumn{1}{|c|}{\multirow{4}{*}{\rotatebox{90}{$\sigma$}}} & \os & \textbf{0.04}          & \textbf{0.08}          & \textbf{0.04}          & \textbf{0.08}          & \textbf{0.04}          & \textbf{0.07}          & \textbf{0.08}          \\ \cline{2-9} 
\multicolumn{1}{|c|}{}                                                             & RA & \textbf{0.04}          & 0.09                   & 0.06                   & 0.09                   & 0.08                   & 0.09                   & 0.11                   \\ \cline{2-9} 
\multicolumn{1}{|c|}{}                                                             & MD & \textbf{0.04}          & 0.09                   & 0.06                   & 0.09                   & 0.07                   & 0.11                   & 0.09                   \\ \cline{2-9} 
\multicolumn{1}{|c|}{}                                                             & DA & \textbf{0.04}          & \textbf{0.08}          & 0.06                   & \textbf{0.08}          & 0.07                   & 0.09                   & 0.09                   \\ \hline
\end{tabular} }
\caption{This table presents the normalized makespan (makespan divided by the theoretical lower bound - lower is better) for cases 1 to 7 for the four compared methods, indicated by: \os~- Constraint-Based Online Scheduling, RA - Random Allocation, MD - Max Duration, DA - Dynamic Allocation. We report mean $\mu$, standard deviation $\sigma$, $10$-th percentile, and $90$-th percentile. As can be seen, \os~outperforms the other methods in all cases. The margin is bigger for the more complex cases where looking ahead makes a difference.}\label{tab:result_table}
\end{table}

A demonstration of applying the methodology to a real robotic system is demonstrated in the accompanying \href{https://youtu.be/YTgaWH19zfA}{video}. The video showcases the robot's responsive behavior towards changes in the workspace or the presence of a human, along with its accurate execution of a series of interconnected tasks.

%The algorithm is robust since the optimal solution was always achieved during the experiments, and there were no conflicts in constraints during rescheduling.

% \section{RSS citations}

% Please make sure to include \verb!natbib.sty! and to use the
% \verb!plainnat.bst! bibliography style. \verb!natbib! provides additional
% citation commands, most usefully \verb!\citet!. For example, rather than the
% awkward construction 

\section{Conclusion} \label{sec:conclusion}

In this paper, we proposed \os, an online scheduling approach for Human-Robot Collaboration based on Constraint Programming based scheduling and behavior trees. 

In our comprehensive evaluation, we compare our proposed online constraint-based scheduling method against two baseline methods and a method from the literature. These comparisons span across seven classes of tasks designed to test different aspects of uncertainty and decision-making flexibility in human-robot collaboration in assembly processes.

To ensure the robustness of our findings, we conducted an extensive set of 56,000 experiments using a probabilistic simulation. This rigorous experimental setup allows us to thoroughly evaluate the performance of our method under a wide range of scenarios and conditions.

Our method significantly outperforms the other methods, particularly in the more complex task classes. This demonstrates its ability to manage uncertainties and enhance human-robot collaboration in assembly tasks.
We gave a preview of our proof-of-concept implementation on a real Franka Emika Panda robot. 

Moreover, our method is highly efficient, running fast enough to be used online and generate new schedules every second. This real-time scheduling capability is crucial for dynamic and uncertain environments, further highlighting the practical applicability of our approach.

The developed tools and a clear baseline enable further research regarding uncertainty-aware scheduling for collaborative manufacturing. Specifically, how to enable  \os~to take known distributions into account, systematic experiments with the real setup, the introduction of more complex tasks, and the learning of probabilistic distributions from experience.

% \section*{Acknowledgments}

% M.I. was supported by the RICAIP project funded by European Union's Horizon 2020 research and innovation program under grant agreement No. 857306

% J.K.B. was supported by the European Regional Development Fund under project Robotics for Industry 4.0 (reg. no. CZ.02.1.01/0.0/0.0/15\_003/0000470).

% \bibliographystyle{plainnat}
% \bibliography{references}

\bibliographystyle{IEEEtran}
\bibliography{references}

\end{document}

%% file: macro.tex
\newcommand{\equationref}[1]{\hyperref[#1]{Eq.~\ref*{#1}}}
\newcommand{\figref}[1]{\hyperref[#1]{Fig.~\ref*{#1}}}
\newcommand{\tabref}[1]{\hyperref[#1]{Table~\ref*{#1}}}
\newcommand{\secref}[1]{\hyperref[#1]{Section~\ref*{#1}}}
\newcommand{\algoref}[1]{\hyperref[#1]{Alg.~\ref*{#1}}}

\newcommand{\etal}[1]{#1 \textit{et al.}}

\newif\ifuseVspace